\documentclass{article}

\PassOptionsToPackage{numbers}{natbib}
\usepackage[final]{neurips_2024_ml4ps}

\usepackage[utf8]{inputenc} 
\usepackage[T1]{fontenc}    
\usepackage{hyperref}       
\usepackage{url}            
\usepackage{booktabs}       
\usepackage{amsfonts}       
\usepackage{nicefrac}       
\usepackage{microtype}      
\usepackage{xcolor}         
\usepackage{graphicx} 
\usepackage{subfig}
\usepackage{pdflscape}

\title{Tomographic SAR Reconstruction for Forest Height Estimation}

\author{%
  Grace Colverd \\
  University of Cambridge \\
  \texttt{gb669@cam.ac.uk}
  \And
   Jumpei Takami \\
  United Nations Office for Outer Space Affairs  \\
  \texttt{jumpei.takami@un.org}
  \And
   Laura Schade \\
UK Department for Energy Security and Net Zero \\
\texttt{laura.schade@energysecurity.gov.uk } \\ 
  \AND
   Karol Bot\\
  INSEC TEC \\
  \And
   Joseph A. Gallego-Mejia\\
Drexel University \\
\texttt{joseph.gallegomejia@drexel.edu}
}

\begin{document}

\maketitle

\begin{abstract}

Tree height estimation serves as an important proxy for biomass estimation in ecological and forestry applications. While traditional methods such as photogrammetry and Light Detection and Ranging (LiDAR) offer accurate height measurements, their application on a global scale is often cost-prohibitive and logistically challenging. In contrast, remote sensing techniques, particularly 3D tomographic reconstruction from Synthetic Aperture Radar (SAR) imagery, provide a scalable solution for global height estimation. SAR images have been used in earth observation contexts due to their ability to work in all weathers, unobscured by clouds. In this study, we use deep learning to estimate forest canopy height directly from 2D Single Look Complex (SLC) images, a derivative of SAR. Our method attempts to bypass traditional tomographic signal processing, potentially reducing latency from SAR capture to end product. We also quantify the impact of varying numbers of SLC images on height estimation accuracy, aiming to inform future satellite operations and optimize data collection strategies. Compared to full tomographic processing combined with deep learning, our minimal method (partial processing + deep learning) falls short, with an error 16-21\% higher, highlighting the continuing relevance of geometric signal processing.

\end{abstract}

\section{Introduction and Methodology}

Estimating biomass is a critical task for assessing global efforts in carbon sequestration, ecosystem conservation, and disaster response readiness, among other environmental objectives. One common approach to biomass estimation is using tree height as a proxy. This can be measured using on-site manual tools or terrestrial Light Detection and Ranging (LiDAR) devices \citep{kwak_detection_2007}. However, these methods are labour-intensive and costly. Remote sensing, particularly through Synthetic Aperture Radar (SAR) imagery, provides a more efficient alternative. SAR is an image captured by an active satellite looking sideways \citep{chan_introduction_2008}. The level 0 SAR or raw SAR image has to be processed using a series of steps such as multiple compressions, motion compensation and calibration, among others. After this series of steps a Single Look Complex (SLC) image is obtained. Each SLC image might be captured using a different incidence angle. According to the slice theorem, a three-dimensional reconstruction can be performed using each SLC. This process involves performing a 1D spectral analysis on co-registered 2D SAR images to capture the vertical reflectivity, i.e., the reflectivity orthogonal to the radar's line of sight \citep{khoshnevis_tutorial_2020}. The challenge then becomes a spectral estimation problem, where the variables to estimate are the vertical reflectivity and the elevation within each pixel. Because the spectral estimation problem is underdetermined, parametric estimation models are required to obtain a solution. Numerous estimation algorithms have been proposed to solve this, including Non-Linear Least Squares, Beamforming \citep{van_veen_beamforming_1988}, Capon \citep{stoica_spectral_2005}, Singular Value Decomposition (SVD), and Wavelet-Based Compressive Sensing \citep{huang_three-dimensional_2017}. Some of these methods build a covariance matrix by comparing a master image with the remaining images and then use this matrix for parametric estimation.
In our work, we leverage the covariance matrix to estimate forest heights, similar to the approach presented in CATSNet \citep{yang_catsnet_2024}. This approach removes the spectral estimation problem and aims to speed up analysis. However, we differ from prior work in that we do not quantise the height estimation and utilise the new TomoSense dataset. 

Additionally, we conduct a detailed study on the impact of using additional SLC images for improving height estimation accuracy, which will be relevant for optimising the future performance of the ESA Biomass Satellite \citep{esa_esa_nodate}. This upcoming mission initiative is designed to operate in two distinct phases: an initial one-year `tomographic mode', during which it will capture 7 SLC images from marginally different incident angles, followed by a three-year `standard mode' operation, capturing 3 SLC images. Quantifying the impacts of different SLC inputs will help optimize the satellite's post-launch performance. 

\subsection{Input Data}
We utilize TomoSense data \citep{tebaldini_tomosense_2023}, which provides aerial SAR data and a LiDAR-based digital terrain model (DTM) and canopy height model (CHM) for a forested region of Eifel Park, Germany. Our analysis uses P-band SAR data (wavelength 69cm, bandwidth 30MHz, resolutions: slant range 5m, azimuth 3m, vertical 3m) which is made up of 28 native SLC images. The data comprises two headings: NW (290°) and SE (110°), representing distinct flight paths and geographic areas. The P band operates in full polarization mode, containing four different polarisation channels: HH, HV, VH and VV. These polarization modes correspond to the polarization that the radar transmits and receives and capture different surface properties such as roughness, shape and material composition \citep{van_zyl_model-based_2011}. We analyse NW and SE headings both separately (as in the original TomoSense report) and together, and analyse the performance of each polarization channel separately. The SLC products have undergone several calibration steps by TomoSense: interferometric calibration for coregistration, polarimetric calibration to ensure consistent features, tomographic calibration for three-dimensional focusing, and ground steering to align ground scattering \citep{tebaldini_tomosense_2023}.   

\subsection{Data Processing}
We further process the TomoSense SLC data to prepare for our CHM modelling pipeline. Our processing includes the following steps. LiDAR forest height scans in original UTM coordinates are transformed into a radar coordinate system congruent with the SLC images. Following the CATSNet methodology \citep{yang_catsnet_2024}, we process the SLC images to generate their covariance matrix. For generating the covariance matrix we are guided by a Tomography Processing Github developed by EO-College\footnote{\url{https://github.com/EO-College/tomography_tutorial}}. We adapted this method to use GPU-aided processing to speed up the generation using CuPy\footnote{\url{https://cupy.dev/}}. This includes the topographic processing based on the provided DTM, and the full covariance matrix generation. From this matrix, we extract three feature vectors: the diagonal, the real elements of the first row, and the complex elements of the first row. These elements represent the variance of each image and the real and imaginary part of the covariance of a master image versus the rest of the images. These are then stacked to create a $3 \times 28$ vector per pixel. This process yields six datasets, each sized $(H,W, 3 \times 28)$, corresponding to different headings and polarizations. For performance evaluation with $n$ input SLC images, we randomly select $n$ images from the data stack (always including the first `master' image) and apply this subset across the three feature vectors, resulting in a stack of size $(H,W,3 \times n)$ where $n \in [3,7,28]$. The processing code will be made available by the corresponding author.

\subsection{Modelling}
The CHM prediction problem is formulated as an image-based regression analysis, predicting the canopy height from the processed SLC image stack per pixel. In our prior works, we achieved success on the TomoSense data with modified three-dimensional convolutional neural networks applied to the 3D TomoSAR data, with U-Net backbones \citep{colverd_3d-sar_2024}. Here, we are using the stack of 2D SLC rather than the 3D TomoSAR, the key difference being that our input data has not undergone the full spectral estimation process used to generate the three-dimensional data from the two-dimensional SLC. We only implement part of the tomographic process, generating the covariance matrix from the SLC but not completing the estimation. We treat the covariance elements as channels, not z-dimensions, and use a 2D U-Net architecture \citep{ronneberger_u-net_2015}. We test using two backbones: a `deep' and `shallow' U-Net (4 vs. 3 encoder/decoder layers). Full model architecture diagrams are given in Appendix \ref{model}. We split training/validation/test data using an adapted quadrant methodology, due to the non-standard shape of the datasets. The quadrant split for heading SE can be seen in the results Figure \ref{fig:error_plot}, where Q1 and Q4 are used for training, Q2a for validation and Q2b for testing. The percentage splits of the data based on input patches for train/val/test are 64/20/16. When training across multiple headings we distribute the train/val/test data across both headings. 

Our methodology includes several key components to ensure robust model training and evaluation. All datasets undergo standardization using a min-max scaler, which is trained on the training data and subsequently applied to the training, validation, and test sets to maintain consistency. We conducted comprehensive hyperparameter sweeps, focusing on learning rate, patch size, and batch size. For model training, we employed the Adam optimizer and utilized Mean Square Error (MSE) as the loss function. To address potential edge effects, we implemented a custom metric to calculate the Mean Absolute Error (MAE) during validation and testing. This metric employs a stride of half the patch size and calculates the error for the central part of the patch, minimising edge effects. Further details on the model training method are given in Appendix \ref{training}.

\subsection{Experiments}
Our study comprises three experiments utilizing different subsets of six derived datasets (two headings × three polarization channels). For each heading/polarization combination, we train separate models targeting CHM. The details of the experiments are given in Table \ref{tab:exps}.

\begin{table}[]
\caption{Experiments}
\label{tab:exps}
\begin{tabular}{@{}lll@{}}
\toprule
\textbf{Name} & \textbf{Experiment}                                                            & \textbf{Description}                                                                                                                                                                          \\ \midrule
1             & \begin{tabular}[c]{@{}l@{}}Heading Performance \\ Comparison\end{tabular}      & \begin{tabular}[c]{@{}l@{}}Compare the performance across the two headings,\\  choose heading for primary study area\end{tabular}                                                             \\
2            & \begin{tabular}[c]{@{}l@{}}Input Optimization: \\ Height Filtered\end{tabular} & \begin{tabular}[c]{@{}l@{}}Quantify the benefit of increasing the number of  SLC images \\ across polarisation channels on a subset of the data: areas >5m \\ based on LiDAR DTM for SE\end{tabular} \\
3          & \begin{tabular}[c]{@{}l@{}}Input Optimisation:\\  Unfiltered Data\end{tabular} & \begin{tabular}[c]{@{}l@{}}Quantify the benefit of increasing the number of SLC images\\  for single-channel on full dataset without height exclusion for SE, \\
 NW and SE+NW. Test impact of SLC image selection\end{tabular}                            \\ \bottomrule
\end{tabular}
\end{table}
Exp 1. will highlight the different qualities of the two heading areas, and provide a choice of heading area to focus on for Exp 2. Exp. 2 is inspired by the TomoSense report, which excluded areas under 5m when predicting height using LiDAR-based canopy maps, removing the impact of soil moisture or ground-level double bounce scattering, a noted issue with SAR data \citep{monteith_temporal_2021}. Exp. 3 recognises that a global model based on satellite data would lack LiDAR scans for height exclusion, and hence uses a full height stack of data. We run this for a single polarisation band across headings SE, NW and SE+NW. Comparisons with prior works require a SLC=28, NW+SE experiment. However, due to memory hardware restrictions, we can only run up to SLC=7 for NW+SE. Hence we test SLC=3,7,28 for single headings (NW and SE) and SLC=3,7 for dual headings (NW+SE).  

\section{Results and Discussion}
\subsection{Experiment 1} 
\begin{figure}[htbp]
    \centering
    \begin{minipage}{0.48\textwidth}
    \caption{Test MAE by Heading, 3 SLC Inputs.}
    \label{fig:heading_compairson}    \includegraphics[width=1\linewidth]{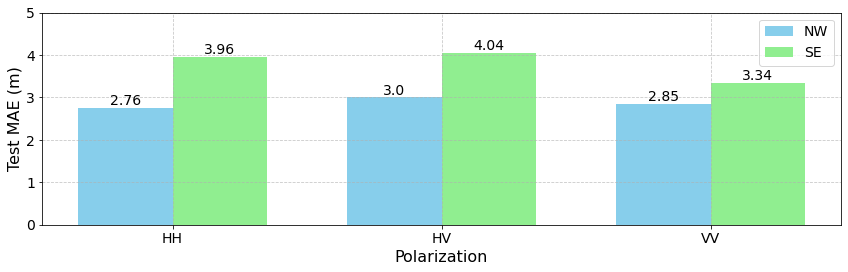}
    \end{minipage}\hfill
    \begin{minipage}{0.48\textwidth}
        \centering
           \caption{Test MAE by SLC and Pol. Channel.}
        \label{fig:test_rmse}
        \includegraphics[width=1
\linewidth]{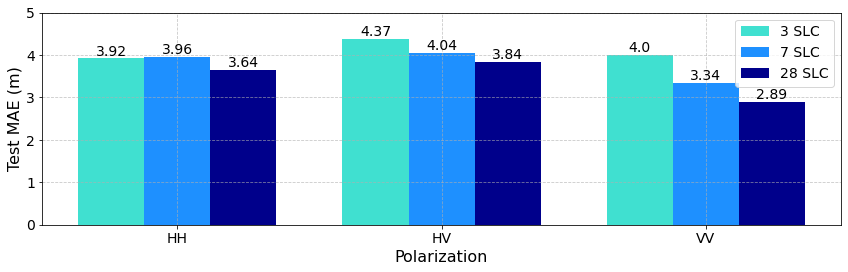}
    \end{minipage}
\end{figure}

The comparison in test MAE performance for headings NW and SE for 3 inputs SLC images is given in Figure \ref{fig:heading_compairson}. Note that the SE dataset seems to have a more challenging test area, with the test MAE greater than for NW. Due to computational constraints, Exp. 2 focuses exclusively on the SE dataset for detailed polarisation / SLC comparisons.  

\subsection{Experiment 2}
The test MAE performance of the different number of input SLC images and polarisation channels is given in Figure \ref{fig:test_rmse}, with the full results table including model settings given in Appendix \ref{results}. The average percentage improvement averaged across the three polarisation channels from 3$\rightarrow$7, and 7$\rightarrow$28 are  8\%, and 9\% for Test MAE respectively.

\begin{figure}[htbp]
\centering
\begin{minipage}[t]{0.48\textwidth}
    \centering
        \caption{CHM Reconstruction L- Prediction, R- Ground truth}
    \label{fig:error_plot}
    \includegraphics[width=\linewidth]{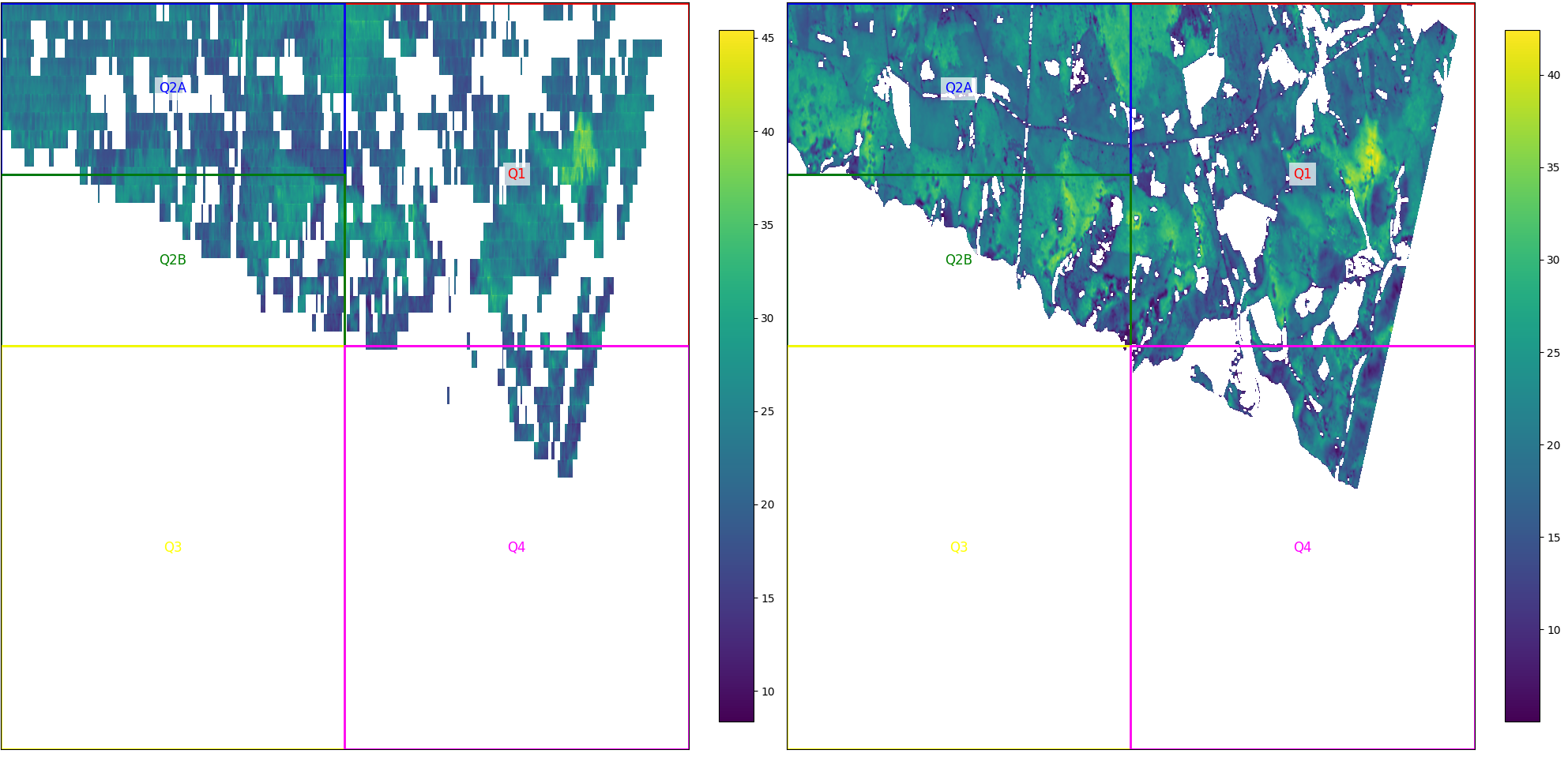}

\end{minipage}%
\hfill
\begin{minipage}[t]{0.48\textwidth}
    \centering
        \caption{Exp. 3 Full data results, P-band, VV}
    \label{tab:full_data}

\begin{tabular}{@{}llllll@{}}
\toprule
Heading                                                                                     & \begin{tabular}[c]{@{}l@{}}Num\\ SLC\end{tabular} & \begin{tabular}[c]{@{}l@{}}Val \\ MAE\end{tabular} & \begin{tabular}[c]{@{}l@{}}Test \\ MAE\end{tabular} & \begin{tabular}[c]{@{}l@{}}Test\\ RMSE\end{tabular} & \begin{tabular}[c]{@{}l@{}}Test \\ R$^2$\end{tabular} \\ \midrule
SE                                                                                          & 3                                                 & 4.22                                               & 5.51                                                & 7.01                                                & 0.17                                                                 \\
SE                                                                                          & 7                                                 & 4.11                                               & 4.59                                                & 6.14                                                & 0.33                                                                 \\
SE                                                                                          & 28                                                & 3.96                                               & 3.87                                                & 5.62                                                & 0.46                                                                 \\
NW                                                                                          & 3                                                 & 4.80                                               & 4.22                                                & 5.96                                                & 0.26                                                                 \\
NW                                                                                          & 7                                                 & 4.23                                               & 3.57                                                & 5.42                                                & 0.37                                                                 \\
NW                                                                                          & 28                                                & 4.21                                               & 3.21                                                & 5.32                                                & 0.39                                                                 \\
SE+NW                                                                                       & 3                                                 & 4.54                                               & 4.42                                                & 5.98                                                & 0.32                                                                 \\
SE+NW                                                                                       & 7                                                 & 4.33                                               & 4.17                                                & 5.71                                                & 0.38                                                                 \\ \midrule
\begin{tabular}[c]{@{}l@{}}SE+NW \\ \citep{colverd_3d-sar_2024} \end{tabular} & 28                                                & 2.98                                               & 3.44                                                & 4.96                                                &                                                                      \\ \bottomrule
\end{tabular}

\end{minipage}
\end{figure}

The VV polarisation channel (vertical transmit, vertical receive) has an overall better ability to discern forest height, across all three SLC inputs. The reduction in test MAE from 3 to 7 SLC for VV is 16.5\%. The top performance by VV is to be expected given VV sensitivity to vertical structures like tree trunks and echoes results seen for L-band SAR data \citep{shimada_tree_2001} and C-band \citep{morrison_ground-based_2013}. Hence VV is selected for Exp 3. The forest reconstruction for the VV channel is given in Figure \ref{fig:error_plot}.

\subsection{Experiment 3}
Figure \ref{tab:full_data} presents the results from CHM prediction using varying numbers of SLC images, without height exclusion for the VV polarisation, across multiple headings. We add results from our prior study using TomoSense 3D TomoSAR input for comparison. The full data results show a similar trend to VV in Exp 2, with a mean reduction in test MAE for 3$\rightarrow$7, and 7$\rightarrow$28 of 16\% and 13\% for NW and SE. Including the data for <5m in this experiment increases the error significantly from 3.34m MAE in Exp 2. to 4.59m MAE for SE heading, 7 SLC inputs, indicating the model struggles to discern the lower portions of the canopy e.g. the river areas and lower canopy.

Given the noted limitations in generating a SE+NW SLC=28 result\footnote{See appendix for hardware details}, we use the single headings performance changes from 3$\rightarrow$7$\rightarrow$28 SLC to temper our expectations of what a SE+NW, SLC=28 result might look like. Assuming the same ratio of performance\footnote{Error reduction from 3$\rightarrow$7 for SE+NW is 5.6\%, 35\% of the mean change 3$\rightarrow$7 for a single heading} comparing the mean of single headings to SE+NW, we expect up to a 5\% decrease in error 7$\rightarrow$28, leading to an estimated test MAE of between 3.98-4.17m (using SLC=7 as the upper bound). Comparing this result to \citep{colverd_3d-sar_2024} this equals a 16 - 21\% increase in MAE for our partial processing method. Given the similarity in prediction pipelines for both works (deep learning with U-NET backbones), this change in performance can be attributed to the input pre-processing (spectral estimation vs. covariance matrix extraction). We conclude that the covariance method with feature extraction is not comparable to full tomographic processing in its current form. Given the problems with the ground layer identified above, it is likely that the full tomographic processing shows the most benefit in reducing the impact of the ground scattering noise. Future work should investigate whether including more of the covariance matrix or augmenting with additional geometrical parameters from the SAR data might improve pre-processing without needing to do full spectral estimation.

\section{Conclusions}
Forest height estimation serves as a proxy for biomass estimation, used for carbon sequestration monitoring and ecosystem management. While remote sensing with compressive sensing has been the traditional approach for SAR data analysis, deep learning now offers new possibilities for improved speed and accuracy. In our study, we investigate the performance of deep learning in predicting forest canopy height  (CHM) from SAR SLC (single-look complex) images, using a case study of SAR data captured from an aerial survey of a forest in Germany. 

We are motivated by the upcoming ESA Biomass Satellite launch in 2025. The new satellite will be a P-Band satellite (compared to prior satellites operating in L-band) and operate in `full tomographic mode' for one year, capturing 7 SLC images per pass. Subsequent years will operate in `standard mode', capturing 3 images per pass (SAR remote sensing requires multiple images (`passes') over a location to make accurate predictions). Hence we are interested in the expected accuracy gains in 7 vs. 3 input images. In addition, traditional tomographic processing requires numerical estimation to solve the spectral estimation problem. We are interested in replacing this with deep learning. 

Our study had two main aims:  1. quantify the expected accuracy gains from increasing the number of input SLC images for SAR data in the P-band, and 2. quantify the performance change by replacing part of the traditional SAR tomography pipeline with deep learning. Our model pipeline takes as input a SLC data stack, from the publicly available TomoSense Data. Instead of applying full tomographic processing (spectral estimation) to the SLC stack, we apply partial SAR processing and deep learning (in the form of a covariance matrix, feature extraction and then a customised two-dimensional U-NET architecture) for canopy height prediction.  We test various inputs into this pipeline. 

Our results comparing the input number of SLC inputs show a 16-17\% reduction in CHM test MAE for 7 vs. 3 images, indicating the first year of satellite operation will have lower error than prior launches. When examining performance for polarisation channels, we find that VV polarisation is the best at CHM prediction, aligning with our expectations and previous works. The best-performing model on the full stack of data achieves a test MAE of 4.17m using 7 SLC images and VV polarization (vertical transmit, vertical receive) for P-band data with 3m vertical resolution. 

When excluding ground-level effects (by filtering to regions with greater than 5m of height based on ground truth), we achieve a reduction of 27\% in test MAE for 7 SLC images, indicating that the model struggles to distinguish genuine scattering from the upper canopy and the double bounce from ground level. Comparisons to our prior work that uses data that received full tomographic processing and deep learning show using our partial processing method increases the MAE by 16-21\% \cite{colverd_3d-sar_2024}. We conclude that the covariance method with feature extraction is not comparable to full tomographic processing in its current form. Future work should integrate other geometric parameters or hyperspectral data to improve the ability of the model to discern the lower canopy and river areas. 

\ack 
This work has been enabled by FDL Europe | Earth Systems Lab (\url{https://fdleurope.org}) a public/private partnership between the European Space Agency (ESA), Trillium Technologies, the University of Oxford and leaders in commercial AI supported by Google Cloud, Scan AI and Nvidia Corporation. 

FDL Europe | Earth Systems Lab and its outputs have been designed, managed and delivered by Trillium Technologies Ltd (\url{trillium.tech}). Trillium Technologies is a research and development company with a focus on intelligent systems and collaborative communities for planetary stewardship, space exploration and human health. 

We express our gratitude to Google Cloud for providing extensive computational resources. The material is based upon work under a programme of, and funded by, the European Space Agency. Any opinions, findings, and conclusions or recommendations expressed in this material are those of the author(s) and do not necessarily reflect the views of the European Space Agency.

FDL Europe | Earth Systems Lab’s public/private partnership ensures that the latest tools and techniques in Artificial Intelligence (AI) and Machine Learning (ML) are applied to basic research priorities for planetary stewardship and disaster response, for all Humankind.

\bibliographystyle{unsrtnat}
\bibliography{references}


\appendix

\section{Appendix}

\subsection{Model Architecture }\label{model}
\begin{figure}
    \centering
        \caption{Model Architecture}
    \label{fig:model-arch}
    \includegraphics[width=0.5\linewidth]{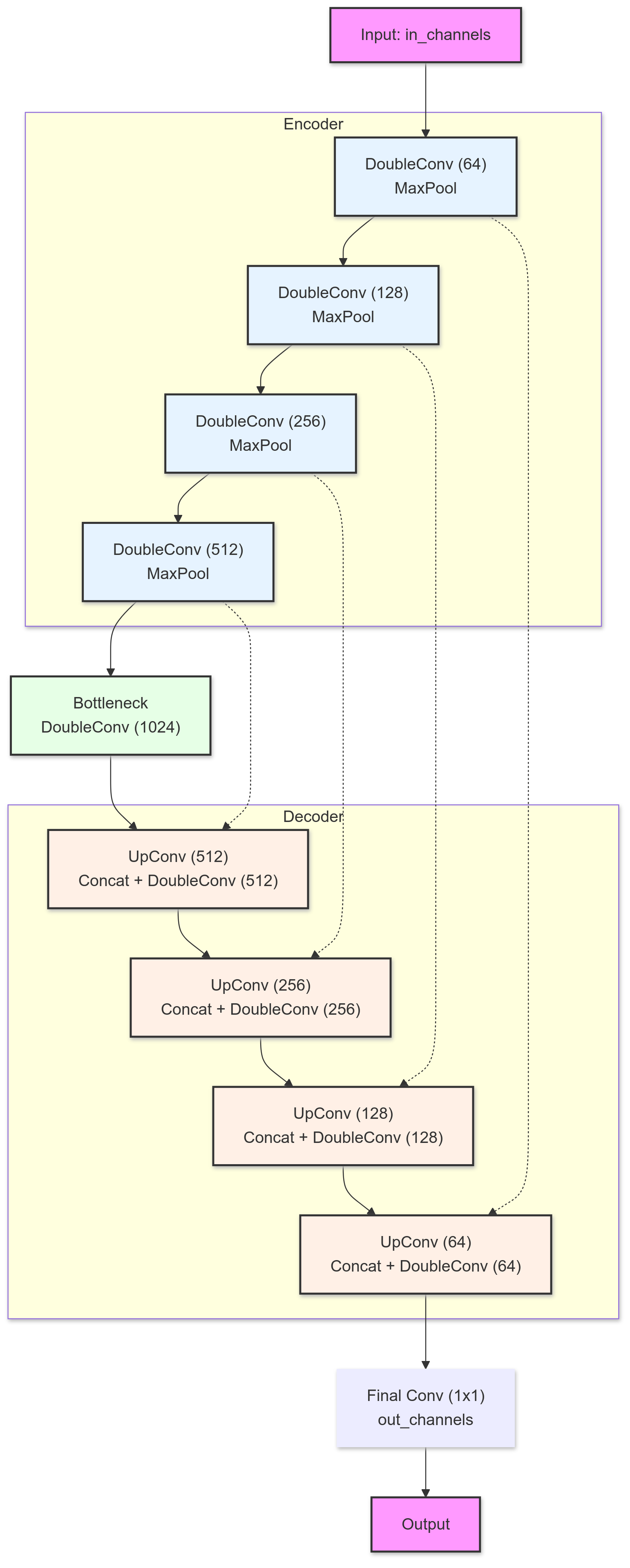}
\end{figure}

\begin{figure}
    \centering
        \caption{Double Convolution Block}
    \label{fig:db_block}
    \includegraphics[width=0.3\linewidth]{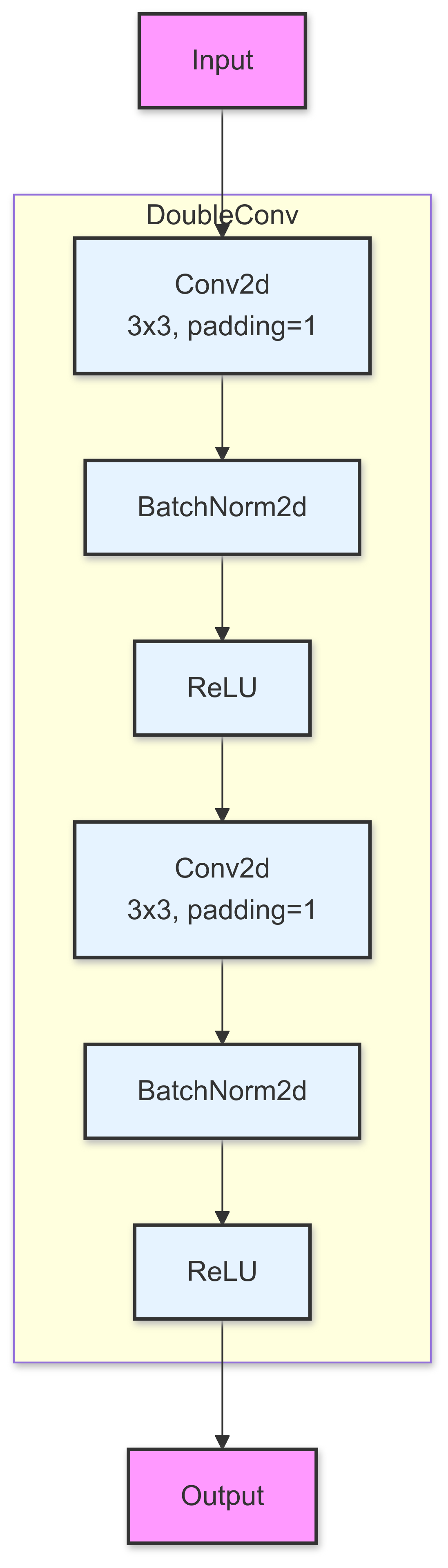}

\end{figure}

The deep model architecture is given in Figure \ref{fig:model-arch}. The double convolution block `DoubleConv' is given in Figure \ref{fig:db_block}. The shallow model removes the final encoder block. Similar performance was observed for both architectures, with the deep method more commonly selected for Exp 3. The shallow version of the model removes the final layer of both the encoder and decoder. 

\subsection{Model Training}\label{training}
Our training process incorporates early stopping based on validation MAE, with an epoch limit of 150 for hyperparameter sweeps (due to early stopping all train for much fewer epochs). When selecting subsets of SLC, we select randomly based on a random seed of 42. 

\subsection{Hardware and Software}
All datasets and experiments were coded in PyTorch. Experiments were run using CuPY, PyTorch and an NVIDIA Tesla T4 GPU with 16GB memory. The GPU ran with NVIDIA driver version 535.183.01 and CUDA version 12.2.  Model training. Experiments and hyperparameter sweeps were run with Weights and Biases (WandB) software. Model training with the subset of the dataset was of the order ~10 minutes, and at least 12 hours were allocated for parameter sweeps for each experiment. Model training with the full dataset was of the order of ~40 minutes and at least 12 hours were allocated to parameter sweeps.

\subsection{Further Results }\label{results}

\subsubsection{Exp. 1 Full Results}
The full model results for Exp 1 are given in table \ref{tab:heading_comp}. These results are for P-Band, with 3 Inputs SLC images, using data with the height filter applied, model sitting and input selection using random seed 42.

\begin{table}[]
\centering 
\caption{Heading Comparison Model results }
\label{tab:heading_comp}
\begin{tabular}{@{}lllllll@{}}
\toprule
Band & Heading & Num SLC & Pol. CH & Val Mae & Test MAE & Test RMSE \\ \midrule
P    & NW      & 7       & HH      & 3.40    & 2.76     & 3.80      \\
P    & NW      & 7       & HV      & 3.23    & 3.00     & 4.05      \\
P    & NW      & 7       & VV      & 3.33    & 2.85     & 3.84      \\
P    & SE      & 7       & HH      & 3.00    & 3.96     & 5.02      \\
P    & SE      & 7       & HV      & 3.23    & 4.04     & 5.27      \\
P    & SE      & 7       & VV      & 2.89    & 3.34     & 4.43      \\ \bottomrule
\end{tabular}
\end{table}

\subsubsection{Exp. 2 Full Results  }

\begin{table}[]
\centering 
\caption{Model results, on datasets with height filter}
\label{tab:full_pol_comp}
\begin{tabular}{@{}llllllllll@{}}
\toprule
\begin{tabular}[c]{@{}l@{}}N\\ SLC\end{tabular} & Pol. CH & \begin{tabular}[c]{@{}l@{}}Model \\ Name\end{tabular} & \begin{tabular}[c]{@{}l@{}}Val \\ Mae\end{tabular} & \begin{tabular}[c]{@{}l@{}}Test \\ MAE\end{tabular} & \begin{tabular}[c]{@{}l@{}}Test \\ RMSE\end{tabular} & \begin{tabular}[c]{@{}l@{}}Test \\ R2\end{tabular} & LR          & \begin{tabular}[c]{@{}l@{}}Batch \\ Size\end{tabular} & \begin{tabular}[c]{@{}l@{}}Patch \\ Size\end{tabular} \\ \midrule
3                                               & HH      & Shallow                                               & 3.04                                               & 3.92                                                & 5.14                                                 & 0.22                                               & 0.00056001  & 16                                                    & (32,32)                                               \\
3                                               & HV      & Shallow                                               & 3.16                                               & 4.37                                                & 5.60                                                 & 0.08                                               & 0.00093618  & 24                                                    & (32,32)                                               \\
3                                               & VV      & Shallow                                               & 3.15                                               & 4.00                                                & 5.16                                                 & 0.22                                               & 0.00059717  & 15                                                    & (32,32)                                               \\
7                                               & HH      & Shallow                                               & 3.00                                               & 3.96                                                & 5.02                                                 & 0.26                                               & 0.0008208   & 20                                                    & (32,32)                                               \\
7                                               & HV      & Shallow                                               & 3.23                                               & 4.04                                                & 5.27                                                 & 0.19                                               & 0.0003316   & 26                                                    & (32,32)                                               \\
7                                               & VV      & Deep                                                  & 2.89                                               & 3.34                                                & 4.43                                                 & 0.42                                               & 0.00069126  & 24                                                    & (32,32)                                               \\
28                                              & HH      & Shallow                                               & 3.11                                               & 3.64                                                & 4.72                                                 & 0.35                                               & 0.000067956 & 38                                                    & (32,32)                                               \\
28                                              & HV      & Shallow                                               & 3.03                                               & 3.84                                                & 4.88                                                 & 0.30                                               & 0.00090485  & 38                                                    & (32,32)                                               \\
28                                              & VV      & Shallow                                               & 2.97                                               & 2.89                                                & 3.97                                                 & 0.54                                               & 0.00057927  & 25                                                    & (32,32)                                               \\ \bottomrule
\end{tabular}
\end{table}

The full polarisation channel results for SE, applying the height threshold (5m) are given in Table \ref{tab:full_pol_comp}.

\subsubsection{Exp 3. Full Results }\label{exp2b_res}
The full results with model settings for Experiment 3 are given in Table \ref{tab:exp2b}. All models use the shallow backbone.

\begin{table}[]
\centering
\caption{Full Model Results for P-Band, SE Heading, VV Chanel}
\label{tab:exp2b}
\begin{tabular}{@{}lllllllll@{}}
\toprule
Heading                                                                                     & \begin{tabular}[c]{@{}l@{}}Num\\ SLC\end{tabular} & \begin{tabular}[c]{@{}l@{}}Val \\ MAE\end{tabular} & \begin{tabular}[c]{@{}l@{}}Test \\ MAE\end{tabular} & \begin{tabular}[c]{@{}l@{}}Test\\ RMSE\end{tabular} & \begin{tabular}[c]{@{}l@{}}Test \\ R$^2$\end{tabular} & \begin{tabular}[c]{@{}l@{}}Learning \\ rate\end{tabular} & \begin{tabular}[c]{@{}l@{}}Batch \\ Size\end{tabular} & \begin{tabular}[c]{@{}l@{}}Patch \\ Size\end{tabular} \\ \midrule
SE                                                                                          & 3                                                 & 4.22                                               & 5.51                                                & 7.01                                                & 0.17                                                                 & 0.00023182                                               & 24                                                    & (64, 64)                                              \\
SE                                                                                          & 7                                                 & 4.11                                               & 4.59                                                & 6.14                                                & 0.33                                                                 & 0.00048592                                               & 21                                                    & (32,32)                                               \\
SE                                                                                          & 28                                                & 3.96                                               & 3.87                                                & 5.62                                                & 0.46                                                                 & 0.00087158                                               & 32                                                    & (64, 64)                                              \\
NW                                                                                          & 3                                                 & 4.80                                               & 4.22                                                & 5.96                                                & 0.26                                                                 & 0.00093345                                               & 29                                                    & (64, 64)                                              \\
NW                                                                                          & 7                                                 & 4.23                                               & 3.57                                                & 5.42                                                & 0.37                                                                 & 0.00079961                                               & 30                                                    & (128,128)                                             \\
NW                                                                                          & 28                                                & 4.21                                               & 3.21                                                & 5.32                                                & 0.39                                                                 & 0.00043169                                               & 22                                                    & (128,128)                                             \\
SE+NW                                                                                       & 3                                                 & 4.54                                               & 4.42                                                & 5.98                                                & 0.32                                                                 & 0.00024355                                               & 19                                                    & (64, 64)                                              \\
SE+NW                                                                                       & 7                                                 & 4.33                                               & 4.17                                                & 5.71                                                & 0.38                                                                 & 0.00087406                                               & 18                                                    & (64, 64)                                              \\ \bottomrule
\end{tabular}
\end{table}

\end{document}